\documentclass[sigconf, 9pt]{acmart}
\AtBeginDocument{% 
    }

\copyrightyear{2024}
\acmYear{2024}
\setcopyright{rightsretained} 
\acmConference[SIGMOD-Companion '25]{Companion of the 2025 International Conference on Management of Data}{June 22--27, 2025}{Berlin,Germany}
\acmBooktitle{Companion of the 2025 International Conference on Management of Data (SIGMOD-Companion '25), June 22--27, 2025, Berlin, Germany}
\acmDOI{10.1145/XXXXX}
\acmISBN{979-8-4007-0422-2/24/06}

\makeatletter

\gdef\@copyrightpermission{
  * Denotes equal first author contribution.
  
  \begin{minipage}{0.3\columnwidth}
   \href{https://creativecommons.org/licenses/by/4.0/}{\includegraphics[width=0.90\textwidth]{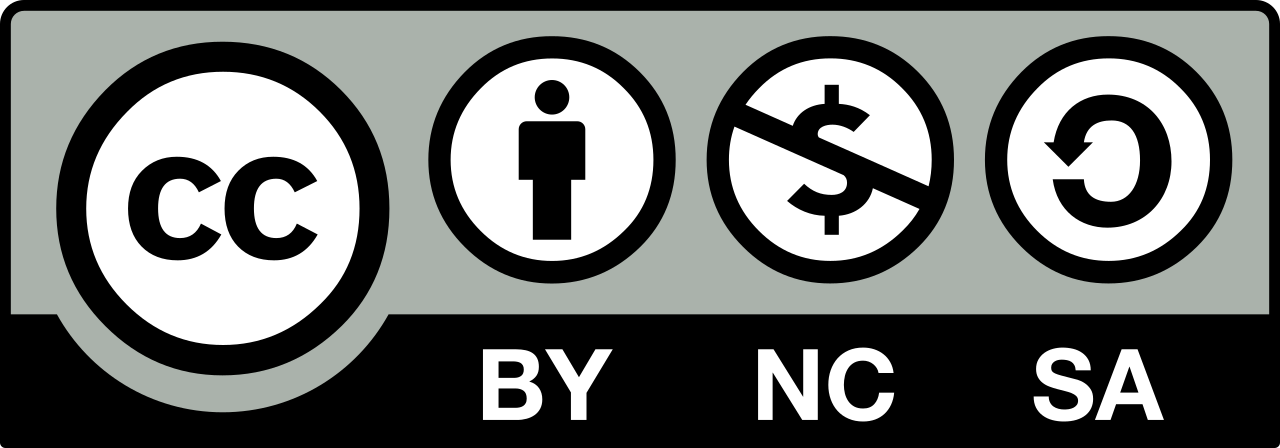}}
  \end{minipage}\hfill
  \begin{minipage}{0.7\columnwidth}
   \href{https://creativecommons.org/licenses/by/4.0/}{This work is licensed under a Creative Commons Attribution International 4.0 License.}
  \end{minipage}
  \vspace{5pt}
}
\makeatother

\usepackage{xcolor}
\usepackage{colortbl}
\usepackage{xspace}
\usepackage{subcaption}
\usepackage{amsmath}
\usepackage{listings}
\usepackage{algorithm}
\usepackage{algpseudocodex}
\usepackage{multirow}
\usepackage{multicol}
\usepackage{tikz}

\usepackage{bbm}
\usepackage{pifont}
\usepackage{hhline}
\usepackage{arydshln}
\usepackage{graphicx}
\usepackage[clock, misc]{ifsym}
\usepackage[inline]{enumitem}
\usepackage[percent]{overpic}
\usepackage{tcolorbox}
\tcbuselibrary{breakable}
\usepackage{tikz}
\usetikzlibrary{arrows.meta, positioning, shapes.geometric}
\usepackage[frozencache,cachedir=.]{minted}
\setminted{
    frame=lines,
    framesep=1mm,
    baselinestretch=1,
    fontsize=\footnotesize,
    linenos,
    numbersep=5pt,
    autogobble,
    breaklines,
    tabsize=2,
    breakautoindent=true,
    mathescape,
}

\let\originalinput\input 

\renewcommand{\input}[1]{%
  \originalinput{#1}
}

\definecolor{darkgreen}{RGB}{0, 100, 0}
\definecolor{experimentblue}{RGB}{127, 159, 186}
\definecolor{experimentgreen}{RGB}{127, 186, 130}

\definecolor{darkorange}{RGB}{240,131,15}

\newcommand{\sys}{\textsc{Palimpzest}\xspace}
\newcommand{\system}{\textsc{Palimpzest}\xspace}
\newcommand{\chat}{\textsc{PalimpChat}\xspace}
\newcommand{\archytas}{\textsc{Archytas}\xspace}

\newcommand{\systemurl}{\url{www.palimpzest.org}\xspace}

\newcommand{\demourl}{\url{www.palimpzest.org/chat}\xspace}

\newcommand{\demovideo}{\url{https://people.csail.mit.edu/chunwei/demo/palimpchat.mp4}\xspace}

\newcommand{\chatcode}{\url{https://github.com/mitdbg/PalimpChat}\xspace}

\begin{document}

\title{\chat{}: Declarative and Interactive AI analytics}
\settopmatter{authorsperrow=3}
\author{Chunwei Liu\textsuperscript{*}}
\affiliation{%
  \institution{MIT CSAIL}
  \city{Cambridge}
  \state{MA}
  \country{USA}
}
\email{chunwei@mit.edu}

\author{Gerardo Vitagliano\textsuperscript{*}}
\affiliation{%
  \institution{MIT CSAIL}
  \city{Cambridge}
  \state{MA}
  \country{USA}
}
\email{gerarvit@mit.edu}

\author{Brandon Rose}
\affiliation{%
  \institution{Jataware}
  % \city{Los Angeles}
  % \state{CA}
  % \postcode{90007}
  \country{USA}
}
\email{brandon@jataware.com}

\author{Matt Prinz}
\affiliation{%
  \institution{Jataware}
  % \city{Los Angeles}
  % \state{CA}
  % \postcode{90007}
  \country{USA}
}
\email{matt@jataware.com}

\author{David Andrew Samson}
\affiliation{%
  \institution{Jataware}
  % \city{Los Angeles}
  % \state{CA}
  % \postcode{90007}
  \country{USA}
}
\email{david@jataware.com}

\author{Michael Cafarella}
\affiliation{%
  \institution{MIT CSAIL}
  \city{Cambridge}
  \state{MA}
  \country{USA}
}
\email{michjc@csail.mit.edu}

% \textsuperscript{*} indicates equal first author contribution.

% %%
% %% By default, the full list of authors will be used in the page
% %% headers. Often, this list is too long, and will overlap
% %% other information printed in the page headers. This command allows
% %% the author to define a more concise list
% %% of authors' names for this purpose.

% \renewcommand{\shortauthors}{Markos Markakis et al.} %% No italics

\begin{abstract}

Thanks to the advances in generative architectures and large language models, data scientists can now code pipelines of AI operations to process large collections of unstructured data.
Recent progress has seen the rise of declarative AI frameworks (e.g., \textsc{palimp}-\textsc{zest}, \textsc{Lotus}, and \textsc{DocETL}) to build optimized and increasingly complex pipelines, but these systems often remain accessible only to expert programmers. 
In this demonstration, we present \chat, a chat-based interface to \sys{} that bridges this gap by letting users create and run sophisticated AI pipelines through natural language alone. By integrating \archytas{}, a ReAct-based reasoning agent, and \sys{}’s suite of relational and LLM-based operators, \chat{} provides a practical illustration of how a chat interface can make declarative AI frameworks truly accessible to non-experts.

Our demo system is publicly available online. At SIGMOD’25, participants can explore three real-world scenarios—scientific discovery, legal discovery, and real estate search—or apply \chat{} to their own datasets. In this paper, we focus on how \chat{}, supported by the \sys{} optimizer, simplifies complex AI workflows such as extracting and analyzing biomedical data. A companion video is also available at \demovideo.

\end{abstract}




\maketitle

% \vspace{-10pt}
\section{Introduction} \label{sec:introduction}

\begin{figure}
    \centering
    \includegraphics[width=\linewidth]{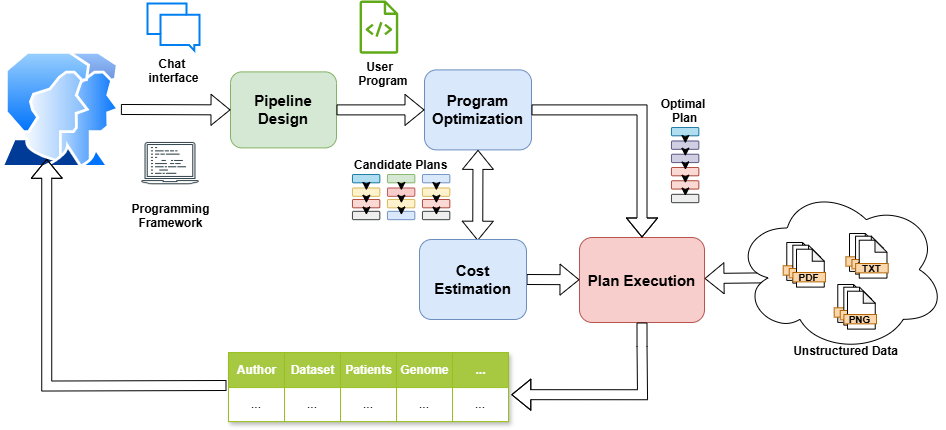}
        \vspace{-2em}
    \caption{An overview of data processing using \chat and \sys}
    \label{fig:architecture-overview}
\end{figure}

Generative AI has transformed our interactions with data and computing by introducing Large Language Models (LLMs) capable of complex tasks such as multimodal extraction, reasoning, and code synthesis~\cite{yao2022react,hong2023metagpt}. 
Yet, implementing such functionality often requires coordinating multiple software stacks—vector databases~\cite{lewis2020retrieval}, relational operators~\cite{palimpzestCIDR}, and novel programming practices like prompt engineering~\cite{dspy}. 
As these AI systems grow in scope, users face major challenges around runtime cost, resource consumption, and the coordination of diverse tools.

\sys{} is one of several new declarative AI systems (e.g., \textsc{Lotus}~\cite{patel2024lotus}, \textsc{DocETL}~\cite{shankar2024docetl}, \textsc{Caesura}~\cite{urban2024caesura}) that adopt a declarative approach for building AI pipelines over unstructured datasets. 
These systems greatly simplify this task for proficient programmers by allowing them to write high-level \emph{logical} specifications - and let automated optimization produce efficient \emph{physical} implementations of these pipelines~\cite{palimpzestCIDR}.
However, this programming model remains inaccessible to less technical users. 
To bridge that gap, we propose \chat{}, a chat-based interface that integrates \sys{} with \archytas{}, a framework to design reasoning agents.
Our system lets users specify data processing pipelines in natural language, while still benefiting from the underlying automated optimization provided by declarative AI frameworks.

In this demonstration, we showcase how both developers and non-experts can collaboratively harness our system. 
All users, regardless of their coding expertise, can design and execute pipelines entirely through \chat{}. 
Furthermore, expert users can either further iterate on the code produced using the chat interface, or program their pipelines directly within \sys{}.
To illustrate the versatility of this approach, we present a real-world use case in scientific discovery, where medical researchers want to identify relevant genomic and proteomic datasets in cancer-related papers. 
Conference attendees can similarly explore additional cases or upload their own unstructured data for processing.

\chat\ is an early, but promising, prototype, and we welcome suggestions for further optimizations and features. 
% We encourage readers to consult our extended technical description in~\cite{palimpzestCIDR}, experiment with our open-source \sys{} code at \systemurl, and interact with \chat{} demo at \demourl and code at \chatcode.
We encourage readers to consult our extended technical description in~\cite{palimpzestCIDR}, experiment with our open-source \sys{} code \footnote{\systemurl}, interact with the \chat{} demo online\footnote{\demourl} or locally \footnote{\chatcode}, or view the video\footnote{\demovideo}.

The remainder of this paper is structured as follows:  
section~\ref{sec:architecture} describes the core architecture of \sys\ and the reasoning framework driving \chat.
Section~\ref{sec:demoscenario} demonstrates a concrete usage scenario through the chat-based interface, and Section~\ref{sec:conclusion} briefly concludes the paper.

\section{System Architecture}
\label{sec:architecture}

The \chat{} system is composed of two interconnected systems: the \sys{} framework, a declarative system to build and automatically optimize data processing pipelines; and the \archytas{} framework, a system to code reasoning agents with access to external tools.
We briefly describe both individually, and then detail how they integrate to form \chat{}.

\subsection{\sys{}}
\sys{} is a declarative system designed to build and optimize pipelines for unstructured data processing, harnessing both LLM-based and conventional operations. At its core, \sys{} programs can be viewed as collections of relational operators. First, users must specify one or more input datasets and their corresponding “Schema.” A schema consists of the attribute names, types, and descriptions used to process the dataset. Given that a dataset can contain unstructured data, \sys{} automatically extracts attribute values from each data record using either LLM-based or conventional methods. Subsequently, users specify a series of transformations over these data records to form the final output. 
% Figure~\ref{fig:operators} highlights the logical operators that \sys{} supports, and Figure~\ref{fig:code} shows a short example program.

Although \sys{} implements most relational algebra operators, our demo emphasizes two special operators: (1) \emph{Convert}, which transforms an object of schema~A into an object of schema~B by computing the fields in~B that do not explicitly exist in~A, and (2) \emph{Filter}, which applies a natural language predicate or UDF on a dataset’s records and returns the subset that satisfies the predicate. All other operations (e.g., \emph{Aggregation}) follow conventional database semantics.
% describe program 

% \subsection{Physical plans}
A \sys{} plan is a sequence of these operators over a dataset. By design, users write \emph{logical} plans only; the choice of the physical implementation is deferred until runtime. For each logical operator, multiple equivalent physical implementations may be available. For instance, a filter operation might be performed via different LLM models, each representing a distinct physical method. More details about logical and physical implementations are available in~\cite{palimpzestCIDR}.

% \subsection{Plan optimization}
After the user specifies their pipelines, 
\sys\ creates a search space of all possible physical plans that implement such plan, which are effectively logically equivalent but may yield outputs of different quality, with a different cost, or with a different runtime.

In a subsequent optimization phase, \sys\ automatically ranks physical plans and selects the most optimal one that meets user-defined preferences.
Users can specify whether they are interested in quality, runtime, or cost of executing their pipelines. 
They may instruct the system to narrow its optimization on one of these dimensions (e.g., to minimize the cost no matter the quality), or specify a meaningful combination of them (e.g., maximize the output quality while being under a certain latency).

% \begin{figure*}
%     \centering
%     \includegraphics[width=1.6\columnwidth]{figures/chat.png}
%     \caption{Chat Interface available at \demourl}
%     \label{fig:schematic}\vspace{-10pt}
% \end{figure*}

\subsection{\archytas{}}
\label{sec:chat}

\begin{figure}[t]
\begin{minted}[escapeinside=||,texcomments]{python}
    @tool()
    async def create_schema(
        self, schema_name: str, schema_description: str, 
        field_names: list, field_descriptions: list, 
        agent: AgentRef) -> str:
        """
        This tool should be used to generate a new extraction schema. 
        The inputs are a schema name and a set of fields. For example, 
        let's say the user is interested in extracting author 
        information from a paper. In this case the schema name might 
        be 'Author' and the fields may be 'name', 'email', 'affiliation'.
        You should provide a short description for each field. 
        Field names cannot have spaces or special characters.
        """
        
        class_name = "{{ schema_name }}"
        schema = {"__doc__": "{{ schema_description }}"}
        for idx, field in enumerate(|\bf{\{\{ field\_names \}\}}|)):
            desc = |\bf{\{\{ field\_descriptions \}\}}|[idx]
            schema[field] = pz.Field(desc=desc)       
        new_schema = type(class_name, (pz.Schema,), schema)
        return new_schema
\end{minted}
\caption{An example Archytas tool used to generate an extraction schema with \sys. Docstrings are important to provide context for the reasoning agent. Inputs variables are injected with the template syntax \texttt{\{\{variable\}\}}.}
\label{fig:toolcode}
\end{figure}

% \subsection{Archytas}
Archytas is a toolbox for enabling LLM agents to interact with various tools in order to solve tasks more effectively, following the ReAct (Reason \& Action) paradigm ~\cite{yao2022react}. 
It is similar in functionality to existing solutions like LangChain~\cite{langchain}, but focuses on providing a streamlined interface for tools. 
By implementing ReAct, an agent can decompose a user request into smaller steps, decide which tools to invoke for each step, provide corresponding input to those tools, and iterate until the task is complete. 
This flexible interface is well-suited for both purely textual tasks and more complex operations (e.g., looking up a dataset, calling AI-based filtering routines, etc.).

\subsection{\chat{}}
The \chat{} interface integrates with \system{} with \archytas{} by exposing a series of tools that the LLM-based agent can leverage. 
Essentially, these tools 
% (shown in Table \ref{tab:agent_tools_short}) 
correspond to templated code snippets that can 1. perform fundamental \sys\ operations (e.g., registering a dataset, generating schemas, filtering records) and 2. orchestrate entire pipelines of transformations.
% without requiring the user to manually write code.
Figure~\ref{fig:toolcode} provides an example of a tool implementation.

\begin{figure}[t]
    \centering
    \includegraphics[width=\linewidth]{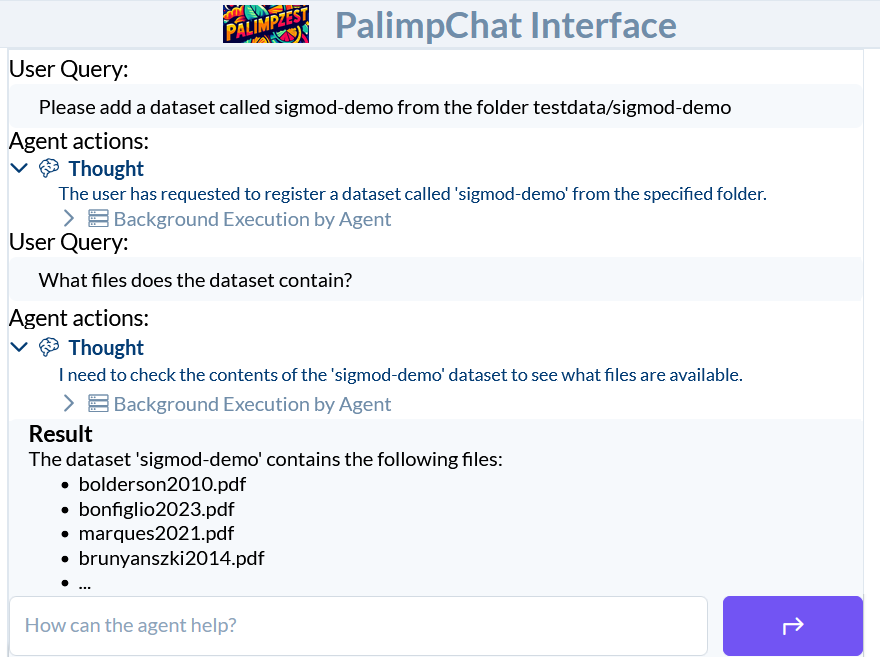}
    \caption{Setting an input dataset through \chat{}}
    \label{fig:dataset_ops}
\end{figure}

The Archytas agent will read tool code as natural language, and consider its doc-string and input/output parameters in order to decide whether to use it to satisfy the user requests.
All tools adhere to a similar pattern in terms of input and output. The general docstring of a tool summarizes what each tool accomplishes and when it is appropriate to use.
The \texttt{Args} section of the docstring can be used to describe the input and output arguments expected for each tool.
Providing a few examples of usage within the docstring proved to be the most efficient solution to improve the quality of the reasoning agent.

The code of each tool is a Python function with the \verb|@tool()| annotation, and a Jinja-based templated syntax can be used to inject run-time variables.
Within the tool code, if a variable is expressed in round brackets as \texttt{\{\{variable\}\}}, the Archytas agent will fill the variable with a variable available at run-time in the Python execution environment.
When parsing a single user request, the reasoning Archytas agent can decide to chain several tool invocations if it deems it necessary to fulfill the desired request (See Section~\ref{sec:demoscenario} for an example).

% \subsection{Beaker}
% Beaker is a custom Jupyter kernel framework designed to integrate seamlessly with web applications. Rather than restricting all interactions to a classic notebook UI, Beaker permits front-end components of an application to leverage long-lived Jupyter sessions in the background.

We host our \chat{} system within a hybrid notebook-chat environment, called Beaker. 
Beaker is an implementation of computational notebooks that integrates AI capabilities into the interactive coding environment. 
Building on Project Jupyter's foundation, Beaker incorporates an AI agent that facilitates code generation and execution while maintaining awareness of the complete notebook state.
This enables context-aware coding, automated debugging, and dependency management, along with comprehensive state management that allows users to restore previous notebook states.

The combination of notebook as well as chat-based features allows users to interact with code, data, and LLM features all from one seamless interface. 
% This powerful combination means a user can, for instance, register a dataset, set a schema, filter, then convert and extract information without ever leaving a single chat or environment.

% While maintaining full compatibility with Jupyter notebook formats, Beaker supports seamless transitions between notebook-style coding and conversational interfaces. 
% Advanced users can customize agents with specific toolsets using the ReAct paradigm, allowing for adaptation to diverse use cases while preserving the familiar notebook interface paradigm.
% \subsection{\sys\ Task Abstraction}
% 
% \chunwei{I think this is a good opportunity to mention something about language-independent abstraction of \sys\ workload. This abstraction could serve as unified representation across any \sys\ language interface.}

\section{Demonstration}
\label{sec:demoscenario}

In this demonstration scenario, we will present a real-world use case leveraging \chat{} to build an end-to-end pipeline for scientific discovery.

\textbf{Use case:} Consider medical researchers interested in performing a literature survey on \emph{colorectal cancer}. 
They are interested in finding studies that report on the correlation between gene mutation and tumor cells, providing or using publicly available datasets of genomic data, and fetching these datasets to run their own analysis.
Assume they have access digital library of scientific papers, but this collection is potentially large, containing unrelated papers, and it is not annotated with metadata about the data sources.

To automate this task, they can build an AI data-processing pipeline using \sys{} through the \chat{} interface.
At a high level, the medical researcher has to upload a collection of papers, filter for those about colorectal cancer, and then extract any reference of publicly available datasets.

% \subsection{Dataset Operations}
% \textbf{Dataset operations:}
At the core of \sys{}, there are \emph{datasets}: collections of input records.
The first step when building a pipeline is to define an input dataset - this could either be a local folder, for which every file will constitute an individual record; or an iterable object in memory, for which every item will be a record.
Additionally, more experienced users can define any custom logic to marshal arbitrary objects or paths into input datasets.

% offers several interfaces for managing datasets, including listing available datasets, registering new ones, and removing unused entries. It also supports retrieving and displaying specific datasets. 

For our demonstration, the user instructs \chat{} to load an input dataset from PDFs of scientific papers contained in a local folder (See Figure~\ref{fig:dataset_ops}).

%\subsection{Data transformations}
The core \chat{} system includes a native \texttt{PDFFile} schema, which is automatically chosen to parse the files in this dataset given their extension. 
However, this schema only represents the filename and the raw textual content extracted for a given paper.

\begin{figure}[t]
    \centering
    \includegraphics[width=\linewidth]{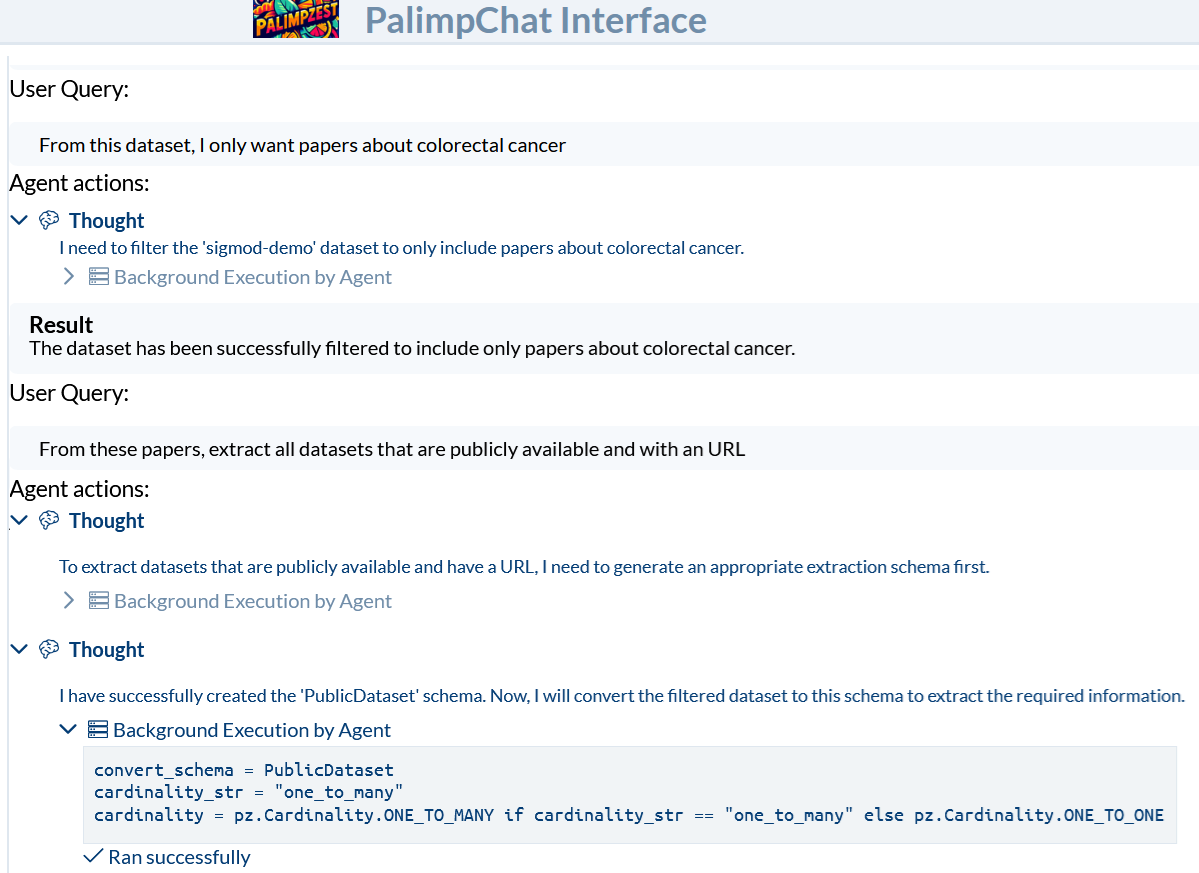}
    \caption{Building a pipeline through natural language. As seen in the last call, the agent reasons and may decide to decompose a user question into several tasks required  before execution.}
    \label{fig:chat-extract}
\end{figure}

Next, the user informs \chat{} that they are interested in papers that are about colorectal cancer, and for these papers, that they would like to extract whatever public dataset is been used by the study in the paper.
From these requests, the reasoning agent creates a pipeline of \sys{} operations: it first uses the \texttt{filter()} function to select papers that satisfy the desired condition.
Then, it dynamically generates an extraction schema with the fields \texttt{dataset\_name}, \texttt{description}, and \texttt{URL}.
Finally, it applies the extraction schema to the filtered dataset with a \texttt{convert()} operation.
Figure~\ref{fig:chat-extract} highlights these steps. 
In the background, \chat{} decomposes the user’s request into sub-steps for data loading, filtering, conversion, and execution.
Before running the full pipeline to process the inputs, the users can specify their optimization goal: minimum cost, maximum quality, or minimum runtime.
\sys{} will automatically determine the appropriate physical plan that implements the logical plan built so far.

The final code generated can be seen in Figure~\ref{fig:finalcode}.

\begin{figure}[t]
    \centering
    \includegraphics[width=\linewidth]{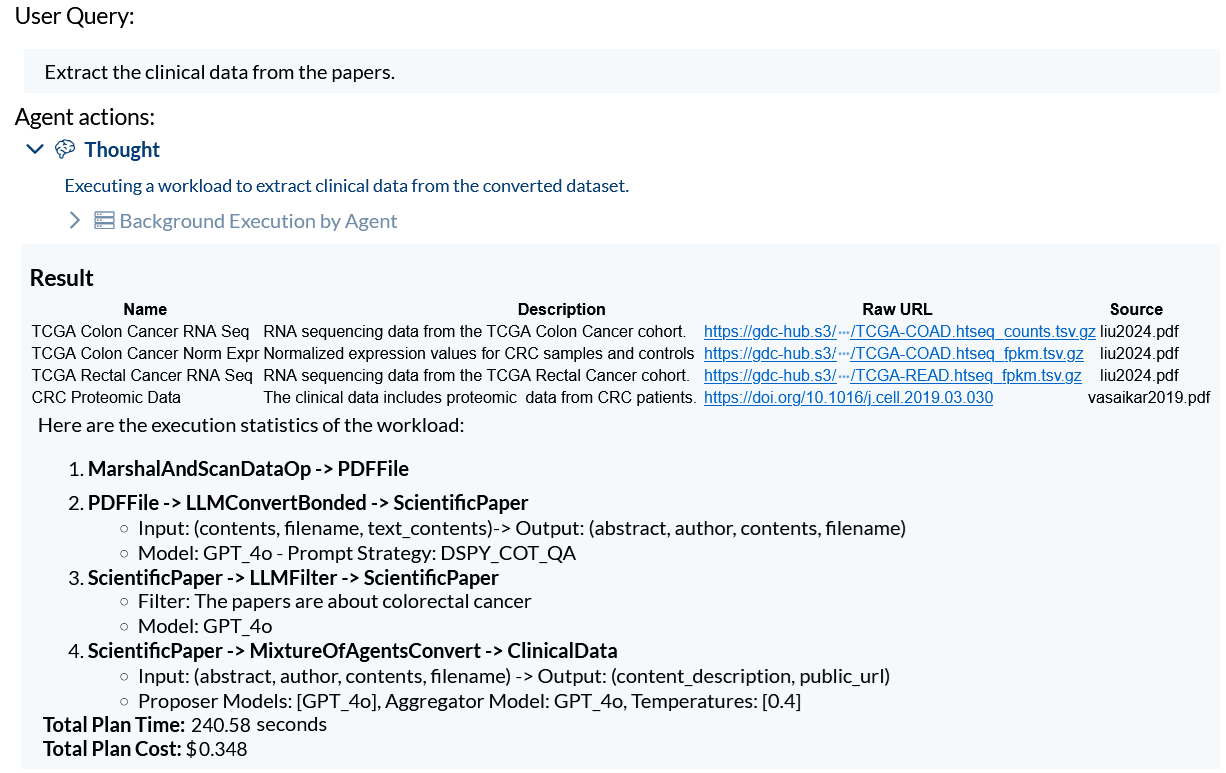}
    \caption{The output of the execution of the scientific discovery use case. Users can visualize both output records, as well as summary information about the plan execution such as the operators chosen and the total pipeline cost and runtime. 
    }
    \label{fig:demo-results}
\end{figure}

When the users prompt the chat to run the pipeline, they will visualize the extracted dataset names and URLs.
Moreover, users can gain insights into the workload execution by asking the system to provide statistics such as how much runtime was needed to produce the output, and how much the LLM invocations costed.
A sample of this output can be seen in Figure~\ref{fig:demo-results}.
For the scientific discovery use case, out of an input dataset of 11 papers, the pipeline managed to extract 6 publicly available datasets related to colorectal cancers, together with the associated URLs.
We manually verified the validity of these URLs.

The execution statistics of the pipeline show that the workload was executed in about 240s and with a cost of about 0.35 USD.

Finally, after building a pipeline, users may continue to iterate on the code produced either through the chat interface or by downloading a Jupyter notebook that contains all inputs and generated snippets of code.

% \begin{figure}[t]
%     \centering
%     \includegraphics[width=\linewidth]{figures/filter_result.png}
%     \caption{\chat{} reasons and decomposes the complex task into compatible steps before execution. 
%     % The outputs of running a \sys{} pipeline. 
%     Users may specify whether they want to optimize for cost, quality, or runtime.
%     }
%     \label{fig:chat-extract}
% \end{figure}

% to create or select a schema that precisely defines the attributes to be extracted or transformed. During this step, users provide a broad schema description—optionally clarifying field requirements, providing attribute descriptions, and indicating which fields are mandatory. If the schema definition is ambiguous or lacks detail, \chat{} prompts the user to supply additional information. This iterative process ensures that \sys{} has a precise understanding of the dataset's structure and context for downstream processing tasks.

% \subsection{Task Execution}
% Finally, we demonstrate how \chat{} orchestrates the user-defined tasks over the dataset. The system interprets the steps needed—such as converting documents to a target schema or filtering records based on user-defined criteria—and executes the pipeline. 

\begin{figure}[t]
\begin{minted}[escapeinside=||,texcomments]{python}
#Set input dataset
schema = PDFFile
dataset = pz.Dataset(source="sigmod-demo", schema=schema)

#Filter dataset
dataset = dataset.filter("The papers are about colorectal cancer")

#Create new schema
class_name = "ClinicalData"
schema = {"__doc__": "A schema for extracting clinical data datasets from papers."}
field_names = ['name', 'description', 'url']
field_descriptions = ['The name of the clinical data dataset', 
    'A short description of the content of the dataset', 
    'The public URL where the dataset can be accessed']
for idx, field enumerate(field_names):
    desc = field_descriptions[idx]
    attributes[name] = pz.Field(desc=desc)
new_class = type(class_name, (pz.Schema,), attributes)

#Perform conversion
convert_schema = ClinicalData
cardinality = pz.Cardinality.ONE_TO_MANY
dataset = dataset.convert(convert_schema, desc=ClinicalData.__doc__, cardinality=cardinality)

#Execute workload
output = dataset
policy = pz.MaxQuality()
records, execution_stats = Execute(output, policy=policy)
\end{minted}
\caption{The final \system{} pipeline built iteratively using the chat interface for the scientific discovery page.}
\label{fig:finalcode}
\end{figure}

\section{Conclusion}
\label{sec:conclusion}
We demonstrate the \chat{} system to interactively build AI pipelines using \sys{} and \archytas{}.
Although our demo did not extensively cover physical optimization aspects, more details can be found in~\cite{palimpzestCIDR}.
The \chat{} interface offers a convenient tool for data practitioners to build complex data processing pipelines with little effort and a soft learning curve.
Our vision is that on the one hand, the future of data engineering will include more and more sophisticated frameworks to build complex applications that mix LLMs and traditional data processing.
On the other hand, tools like \chat{} will assist developers and make it easier to adopt new technologies and programming paradigms.

\begin{acks}
    We gratefully acknowledge the support of the DARPA ASKEM project (Award No. HR00112220042), and the ARPA-H Biomedical Data Fabric project (NSF DBI 2327954).
\end{acks}

\bibliographystyle{ACM-Reference-Format}
\bibliography{bib}

\end{document}